# Exploring Multimodal Large Language Models for Radiology Report Error-checking

Jinge Wu[1,*,#], Yunsoo Kim[1,*,#], Eva C. Keller[1], Jamie Chow[1],

Adam P. Levine[2], Nikolas Pontikos[3,4,5], Zina Ibrahim[6], Paul Taylor[1], Michelle C. Williams[7],

Honghan Wu[1,#]

[1] Institute of Health Informatics, University College London, London, UK

[2] Research Department of Pathology, University College London, London, UK

[3] UCL Genetics Institute, University College London, London, UK

[4] Institute of Ophthalmology, University College London, London, UK

[5] Moorfields Eye Hospital, London, UK

[6] Department of Biostatistics and Health Informatics, King's College London, London, UK

[7] Centre for Cardiovascular Science, University of Edinburgh, Edinburgh, UK

[*]The authors have contributed equally.

[#]Corresponding Author: {jinge.wu.20, yunsoo.kim.23, honghan.wu}@ucl.ac.uk

**Abstract**

**Objective:** This paper proposes one of the first clinical applications of multimodal large language models (LLMs) as an assistant for radiologists to check errors in their reports.

**Methods:** We created a benchmark from three real-world radiology datasets (MIMIC-CXR, IU X-ray, and INSPECT). A subset of original reports was modified to contain synthetic errors by introducing three types of mistakes: "insert", "remove", and "substitute". The evaluation contained two difficulty levels: SIMPLE for binary error-checking and COMPLEX for identifying error types. We evaluated the performance with state-of-the-art multimodal LLMs including LLaVA (Large Language and Visual Assistant) variant models, our instruction-tuned model, and BiomedCLIP. Additionally, a domain expert evaluation was conducted on a small test set.

**Results and Discussion:** At the SIMPLE level, our fine-tuned model significantly enhanced performance by 47.4% and 25.4% on MIMIC-CXR and IU X-ray data, respectively. This performance boost is also observed in unseen modality, CT scans, as the model performed 19.46% better than the baseline model. The model also surpassed the domain expert's accuracy in the MIMIC-CXR dataset by 1.67%. Notably, among the subsets (N=21) of the test set where a clinician did not achieve the correct conclusion, the LLaVA ensemble mode correctly identified 71.4% of these cases. However, all models performed poorly in identifying mistake types, underscoring the difficulty of the COMPLEX level.

**Conclusion:** This study marks a promising step toward utilizing multimodal LLMs to enhance diagnostic accuracy in radiology. The ensemble model demonstrated comparable performance to clinicians, even capturing errors overlooked by humans.

## INTRODUCTION

Clinical reports, encompassing a wide range of medical documentation, are essential in patient care for recording diagnoses, treatments, and patient outcomes. Radiology reports, a critical subset of clinical reports, provide detailed interpretations of medical imaging studies such as X-rays, CT scans, and MRI scans. Their accuracy directly influences clinical diagnosis and subsequent patient care decision-making. The complexity and importance of these reports are highlighted in the medical literature, underscoring the need for precision and clarity in their formulation (1-2). This precision is not only critical for diagnosing acute conditions but also for guiding long-term treatment plans for complex and chronic illnesses such as cancer (3).

The advent of advanced natural language processing (NLP) techniques, notably large language models (LLMs) like the GPT series (4-7), could bring a transformative change in how clinical reports are generated and utilized. These models, capable of automated report generation and summarization, mark a significant leap forward in medical informatics. For example, Zhang et al. (8) proposed a machine-learning method for radiology report summarization tasks. Another study proposed a pre-constructed graph embedding module to assist the generation of reports (9).

More recently, the field has begun to explore multimodal applications, as highlighted in research by Xue et al. (10). This emerging area focuses on integrating visual data, such as radiological images, with textual analysis to generate more comprehensive and accurate clinical reports. This approach exemplifies the growing trend in medical AI to synergize different data modalities for improved patient care and report accuracy.

However, despite their potential, the application of these models, particularly in specialized fields like radiology, comes with significant challenges. It has been noted that original clinical reports authored by clinicians can sometimes contain errors (11). These inaccuracies, though often subtle, can be human-derived and system-derived and would have far-reaching implications. In particular, errors in clinical reports can lead to misinformed decision-making, potentially compromising patient safety (12). Moreover, when such reports are used as training data for models, these errors are inadvertently learned and perpetuated by the model, thus undermining the reliability of the models. Therefore, identifying and rectifying these errors is crucial for ensuring the accuracy and dependability of both the clinical reports and the models trained on them. Another critical challenge arises from the models' reliance on textual data. The recent study introduces a benchmark for error-checking in radiology reports, yet its exclusive focus on the reports themselves deviates significantly from real-world situations (13). Radiology inherently involves complex imaging data, which these text-centric models might struggle to interpret and correlate with clinical contexts accurately. Such gaps in interpretation can lead to serious errors in patient care, including incorrect diagnoses and inappropriate treatment recommendations. Therefore, the adoption of a multimodal approach in report auditing is important, as it bridges the gap between text-based analysis and the nuanced understanding of imaging data, ensuring a more accurate and holistic evaluation of patient information.

To address current challenges in clinical accuracy, we propose an innovative error-checking methodology that combines textual and imaging data (Figure 1). This approach is crucial for improving clinical practice by ensuring the reliability of radiology reports. Our method involves the creation of a unique dataset, where radiology reports are modified to simulate real-world errors. Pinpointing errors or inconsistencies in existing reports with multimodal LLMs is beneficial in assisting clinicians in verifying the quality of reports.

By employing cutting-edge vision-language models (VLMs) for evaluation, our approach establishes a pioneering standard in the evaluation of multimodal clinical tasks. This advancement is instrumental in propelling the application of multimodal language models within specialized domains, such as radiology. Furthermore, to better conduct this task, we fine-tuned the open domain multimodal LLM, LLaVA 1.5 model (20) with a clinical error-checking instruction tuning dataset. The instruction-tuned model showed its effectiveness and improvement on this task, which can better serve as an assistant for radiologists to check for errors in their reports.

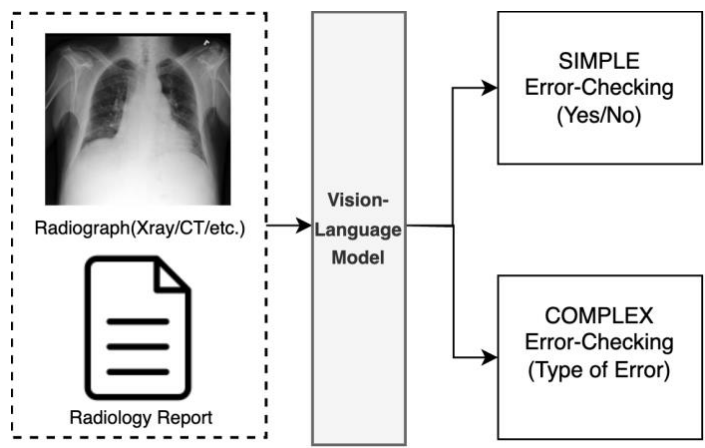

**Figure 1.** Error-checking task description.

**MATERIALS AND METHODS**

**Evaluation data creation**

We use chest radiology studies collected from MIMIC-CXR (14) and IU X-ray (15). The MIMIC-CXR data includes chest radiology reports with chest X-rays performed at the Beth Israel Deaconess Medical Center in Boston, MA between 2011-2016. IU X-ray originates from the Indiana University School of Medicine. Similar to MIMIC-CXR, it contains chest radiology reports with chest X-rays. For each report, we pair it with the image using posterior to anterior (PA) view in IU X-ray and MIMIC-CXR.

To evaluate the generalizability of the models, we further include another dataset Integrating Numerous Sources for Prognostic Evaluation of Clinical Timelines (INSPECT) (25), which contains de-identified longitudinal records from a large cohort of pulmonary embolism (PE) patients with computed tomography pulmonary angiography (CTPA) scans. Due to the partial release of the dataset, with the full data still under review, we utilized the available subset of samples for evaluation. Given the models' limitation to processing only a single image input, each report is associated with the median slice of the CT scan.

We use the entity retrieval toolkit, SemEHR (16), specifically tailored for analyzing clinical reports. This tool is designed to identify clinical concepts embedded within the reports. Following the automated identification process, we asked expert radiologists to select and rank the clinical concepts that are quintessential for radiographic studies. Based on this, the evaluation data for error-checking was created by modifying the original report (**N**) with three types of mistakes on mentions of those top 30 expert-ranked most important concepts:

1) "substitute" mistake (**W**): to randomly replace important concepts with other irrelevant ones. Figure 2 shows the modified report with the "substitute" mistake, where we replaced the original concept *"atelectas"* with *"calcinoses"*.

2) "insert" mistake (**I**): to add an irrelevant sentence to the original report. In Figure 2, there is an example illustrating our evaluation data. For the *"insert"* mistakes, we added a fake sentence *"Hyperemia indicates a foreign body in the right subclavian"*.

3) "remove" mistake (**R**): to randomly remove a sentence with important concepts from the original report. Figure 2 shows that red text with strikethrough shows the removed sentence.

Our evaluation dataset comprises a mix of original and erroneous reports, as outlined in Table 1 which details the data quantity and report lengths. We randomly selected 1,000 reports from both the IU X-ray and MIMIC-CXR datasets. For the CT modality, we gathered 115 reports from INSPECT. Approximately 75% of these are original reports, while the remaining 25% contain various types of errors, with each featuring only one mistake. More detailed statistics are given in

Table 1. This approach aims to mirror real-world scenarios where only a minority of reports are expected to have errors. In terms of report length, IU X-ray reports are, on average, about 100 words shorter than those from MIMIC-CXR, while INSPECT reports are, on average, the longest (435 words) (Table 1).

| Original report | "substitute" mistake | "insert" mistake | "remove" mistake |
|---|---|---|---|
| Single lead pacemaker is unchanged with lead extending to the region of the right ventricle. Again noted, is a large right pleural effusion. Associated compressive atelectasis in the right middle and lower lobes is again seen. Left lung is essentially clear without large effusion or focal consolidation. Mild interstitial edema is present. The heart remains enlarged. No pneumothorax. | Single lead pacemaker is unchanged with lead extending to the region of the right ventricle. Again noted, is a large right pleural effusion. Associated compressive **calcinoses** in the right middle and lower lobes is again seen. Left lung is essentially clear without large effusion or focal consolidation. Mild interstitial edema is present. The heart remains enlarged. No pneumothorax. | Single lead pacemaker is unchanged with lead extending to the region of the right ventricle. Again noted, is a large right pleural effusion. Associated compressive atelectasis in the right middle and lower lobes is again seen. **Hyperemia indicates a foreign body in the right subclavian.** Left lung is essentially clear without large effusion or focal consolidation. Mild interstitial edema is present. The heart remains enlarged. No pneumothorax. | Single lead pacemaker is unchanged with lead extending to the region of the right ventricle. Again noted, is a large right pleural effusion. Associated compressive atelectasis in the right middle and lower lobes is again seen. Left lung is essentially clear without large effusion or focal consolidation. ~~Mild interstitial edema is present.~~ The heart remains enlarged. No pneumothorax. |

**Figure 2.** Example of a report with each type of mistake. Red texts indicate the synthetic mistakes that are introduced in the original report.

Our evaluation dataset comprises a mix of original and erroneous reports, as outlined in Table 1 which details the data quantity and report lengths. We randomly selected 1,000 reports from both the IU X-ray and MIMIC-CXR datasets. For the CT modality, we gathered 115 reports from INSPECT. Approximately 75% of these are original reports, while the remaining 25% contain various types of errors, with each featuring only one mistake. More detailed statistics are given in Table 1. This approach aims to mirror real-world scenarios where only a minority of reports are expected to have errors. In terms of report length, IU X-ray reports are, on average, about 100 words shorter than those from MIMIC-CXR, while INSPECT reports are, on average, the longest (435 words) (Table 1).

**Table 1.** Evaluation data description. The sample size and length of reports are described. N - original report, W - reports with "substitute" mistake, I - reports with "insert" mistake, R - reports with "remove" mistake. The total sample size and average length of reports are provided as well.

| Dataset | Sample size | | | | | Length of reports | | | | |
|---|---|---|---|---|---|---|---|---|---|---|
| | Total. | N | W | I | R | Avg. | N | W | I | R |
| IU X-ray | 1,000 | 743 | 99 | 74 | 84 | 209 | 206 | 233 | 243 | 171 |
| MIMIC-CXR | 1,000 | 756 | 101 | 76 | 67 | 302 | 302 | 294 | 235 | 376 |
| INSPECT | 115 | 87 | 10 | 8 | 10 | 435 | 471 | 261 | 390 | 336 |

**Error-checking problem definition**

The primary task revolves around examining the radiograph alongside its corresponding report to determine the presence of errors. Notably, the challenge extends beyond mere identification to encompass categorization, requiring the model to not only confirm the existence of errors but also classify the specific type of error or discrepancy within the report. The evaluation framework is structured with two levels of difficulty:

**SIMPLE level**

The model is tasked with determining if there is any error between the input report and the corresponding image. The expected responses are binary, with "Y" indicating the presence of inconsistency and "N" indicating no inconsistency.

**COMPLEX level**

Beyond merely identifying the existence of error, the model in this setting also categorizes the type of error detected. The categorization includes "W" for instances where an important concept was replaced, "R" for the removal of a phrase or sentence, and "I" for the insertion of a phrase or sentence.

This two-level design aims to comprehensively evaluate the model's ability in error-checking, from simple binary assessments to nuanced identification and categorization of different types of mistakes in radiological reports.

**SHARED SYSTEM MESSAGE**

You are a senior radiologist and expert in chest xray report. Always answer accurately and precisely. If a question does not make any sense, or is not factually coherent, explain why instead of answering something not correct. If you don't know the answer to a question, please don't share false information. Below is an instruction that describes a task, paired with an input that provides further context. Write a response that appropriately completes the request.

| SIMPLE | COMPLEX |
|---|---|
| ### Instruction:<br>Your junior radiologist made a report with the following findings about the chest xray image. You need to verify if there is any mistakes or fact inconsistency in findings. Answer with the option's letter from the given choices directly: 'N' - no mistakes or no fact inconsistency in findings, 'Y' - mistakes or fact inconsistency in findings.<br><br>### Findings:<br>{findings}<br><br>### Response: | ### Instruction:<br>Your junior radiologist made a report with the following findings about the chest xray image. You need to verify if there is any mistakes or fact inconsistency in findings. Answer with the option's letter from the given choices directly: 'N' - no mistakes or no fact inconsistency in findings, 'W' - used a wrong word in findings, 'I' - inserted a irrelevant sentence in findings, 'R' - removed a relevant sentence in findings.<br><br>### Findings:<br>{findings}<br><br>### Response: |

**Figure 3.** SIMPLE and COMPLEX level prompt template.

**Models**

LLaVA is the first model that used a visual instruction dataset to train a LLM into a VLM (19). We selected the following three LLaVA models as baseline models and compared the performance of the evaluation datasets with our instruction-tuned model. We further include BiomedCLIP for comparison to comprehend the study of multimodal LLMs with different architectures (26).

**LLaVA-0:** LLaVA is one of the pioneering VLMs. The model weight we used is the early version of LLaVA that combines a vision encoder and Vicuna, which is LLaMA trained for instruction purposes, for general-purpose visual and language understanding (17). The model is trained with visual instruction tuning techniques, demonstrating its powerfulness when presented with unseen images and instructions. We will call this model LLaVA-0 hereafter. We used the 7B model.

**LLaVA-Med:** LLaVA-Med is one of the first VLMs in the biomedical domain (21). LLaVA-Med is initialized with LLaVA-0 and then continuously trained with a comprehensive dataset of biomedical figure-caption pairs sourced from PubMed Central. Only the 7B model is available publicly.

**LLaVA-1.5:** LLaVA-1.5 is a general domain VLM that uses the LLaMA2 model, which has a significant improvement in language understanding when compared with LLaMA, as the backbone LLM (18,20). There are two significant improvements besides the change of the backbone LLM. Firstly, the addition of an MLP vision-language connector enhanced the system's capabilities. Secondly, the integration of academic task-oriented data further enhanced its performance and effectiveness. LLaVA-1.5 is available in 2 model sizes, 7B and 13B models, and we used both models.

**LLaVA-CXR and LLaVA-CXR-A:** For the fine-tuning of LLaVA-1.5 7B with the MIMIC-CXR error-checking instruction tuning dataset, we trained two versions of the model: LLaVA-CXR and LLaVA-CXR-A. For both models, we constructed the instruction tuning dataset for training using 17,000 MIMIC-CXR X-rays and reports including 8,492 reports with mistakes and 8,508 reports without mistakes. This is in the same way as the evaluation dataset, except we included the labels for model training. For LLaVA-CXR, we used the SIMPLE setting (binary checking) for training. For LLaVA-CXR-A, we used the COMPLEX setting (multi-class checking) for training.

The fine-tuning code was obtained from LLaVA's official GitHub page (20). Due to GPU resource limitation, we used several techniques such as LoRA, ZeRO3, and Flash Attention to reduce the GPU memory requirement (22-24). We used the same hyperparameter settings as per the LLaVA guidance on GitHub for task-specific fine-tuning, except for reducing the training batch size to 8 and the number of epochs to 2.

**BiomedCLIP:** BiomedCLIP is a biomedical VLM that is pretrained on PMC-15M, a dataset of 15 million figure-caption pairs extracted from biomedical research articles in PubMed Central, using contrastive learning (26). In terms of the architecture, it uses PubMedBERT as the text encoder and Vision Transformer as the image encoder, with domain-specific adaptations.

**Prompting strategy**

In leveraging the LLaVA models, we explored two prompting strategies: zero-shot and in-context learning (few-shot), recognizing the pivotal role that these strategies play in shaping the models' responses. By utilizing zero-shot learning, we sought to assess the model's innate understanding of radiographs and reports by generating accurate responses without the provision of specific examples (shots). Zero-shot learning prompt template is shown in Figure 3. In-context learning, which is widely accepted to improve the performance of LLMs, was conducted to assess the model's learning ability based on provided examples.

**Evaluation**

For evaluation, the temperature and the number of output tokens were constrained to 0 and 1 respectively. This setting ensures consistent and predictable outputs. The evaluation metric is F1 for both levels of difficulty. For the calculation of the F1 score, we treated reports with mistakes as positive regardless of the type of the mistake.

For few-shot prompting, we selected 200 samples for each dataset from the whole dataset and made sure there was no overlap between these 200 samples and our evaluation dataset of 1000 samples. As the number of INSPECT dataset is smaller than 200 samples, we excluded the INSPECT dataset for the in-context learning evaluation. Also, we did not include BiomedCLIP for in-context learning evaluation due to its' smaller maximum sequence length, 512, which limited inputting the examples of the findings report.

**Human Evaluation**

Two clinicians evaluated a random selection of 60 samples drawn from the MIMIC and IU X-ray datasets, with 30 samples from each. Within the dataset, 16 samples were selected for each type of mistake, totaling 48 mistakes, while the remaining 12 reports were 'original.' Given the change in the dataset's mistake proportion, accuracy was chosen as the metric instead of the F1 score. We also limited the difficulty of the task to be SIMPLE.

**RESULTS**

**Zero-shot Evaluation**

Table 2 shows that instruction tuning within the medical domain enhanced the LLaVA variant models' inherent ability to perform the error-checking task in the SIMPLE setting. Specifically, LLaVA-Med was better than its backbone model LLaVA-0 by 20.61 in MIMIC-CXR, by 21.03 in IU X-ray, and by 13.28 in INSPECT. Likewise, the performance of LLaVA-CXR surpassed its baseline model, LLaVA-1.5-7B, by 47.39 in MIMIC-CXR, 25.40 in IU X-ray, and 19.46 in INSPECT, positioning LLaVA-CXR as the top-performing model for the SIMPLE difficulty level.

**Table 2.** Zero-shot Evaluation Result (F1-score). The best result is bolded for each task.

|  | MIMIC-CXR | | IU X-ray | | INSPECT | |
| --- | --- | --- | --- | --- | --- | --- |
| Model | SIMPLE | COMPLEX | SIMPLE | COMPLEX | SIMPLE | COMPLEX |
| BiomedCLIP | 33.52 | **12.04** | 38.48 | 12.84 | 32.06 | **14.63** |
| LLaVA-1.5-13B | 0.00 | 0.00 | 0.00 | 0.00 | 0.00 | 0.00 |
| LLaVA-0 | 12.91 | 3.36 | 8.61 | 0.00 | 6.72 | 0.00 |
| LLaVA-Med | 33.52 | 0.40 | 29.64 | 4.91 | 20.00 | 13.68 |
| LLaVA-1.5-7B | 35.79 | 0.00 | 40.60 | 0.00 | 35.71 | 0.00 |
| LLaVA-CXR-A | 39.16 | 11.27 | 32.36 | **15.66** | 26.83 | 14.08 |
| **LLaVA-CXR** | **83.18** | 0.00 | **66.00** | 0.00 | **55.17** | 0.00 |

However, all the models, including the fine-tuned models, performed poorly in the COMPLEX setting. Despite this, among LLaVA models, our fine-tuned model for both SIMPLE and COMPLEX settings, LLaVA-CXR-A, was the best-performing model for all three datasets.

Although larger models are known to exhibit superior performance, our evaluation of LLaVA-1.5-13B shows the contrary. As it answered "N" all the time, it performed the worst in all settings and datasets, suggesting further study is needed for the relationship between model size and performance of the model in the error-checking task. It is also noteworthy to mention that the previous medical VLM, BiomedCLIP, performed reasonably well on the tasks. Still, our LLaVA-CXR performed better than BiomedCLIP in the SIMPLE setting. Also, LLaVA-CXR-A performed better in the IU X-ray dataset in the COMPLEX setting and fell slightly behind BiomedCLIP in other datasets.

**Findings with "remove" mistake**

Single lead pacemaker is unchanged with lead extending to the region of the right ventricle. Again noted, is a large right pleural effusion. Associated compressive atelectasis in the right middle and lower lobes is again seen. Left lung is essentially clear without large effusion or focal consolidation. ~~Mild interstitial edema is present.~~ The heart remains enlarged. No pneumothorax.

| LLaVA-0 | | LLaVA-Med | | LLaVA-1.5-7B | | LLaVA-1.5-13B | |
|---|---|---|---|---|---|---|---|
| SIMPLE | COMPLEX | SIMPLE | COMPLEX | SIMPLE | COMPLEX | SIMPLE | COMPLEX |
| "Based" | "Based" | "Based" | "I" | "Y" | "N" | "N" | "N" |

| BiomedCLIP | | LLaVA-CXR-A | | LLaVA-CXR | | Ground Truth | |
|---|---|---|---|---|---|---|---|
| SIMPLE | COMPLEX | SIMPLE | COMPLEX | SIMPLE | COMPLEX | SIMPLE | COMPLEX |
| "N" | "R" | "N" | "N" | "Y" | "Y" | "Y" | "R" |

**Figure 4.** Qualitative analysis of one sample. The result is sampled from the zero-shot evaluation result.

Figure 4 shows an example of a "remove" mistake and the responses from all models. Both LLaVA-0 and LLaVA-Med gave a response of "Based", which is irrelevant given the prompt (as detailed in Figure 3). The example also shows that all the models except BiomedCLIP got a wrong answer for the COMPLEX difficulty, while LLaVA-1.5-7B and LLaVA-CXR yielded correct responses for the SIMPLE difficulty. It's worth noting that LLaVA-CXR, while also producing an irrelevant answer, may be constrained by the instruction tuning dataset we used for training, which only allows for a binary "Y" or "N" response.

**Exploration of In-context Learning**

**Table 3.** In-context learning Evaluation Result. 2-Shot, and 4-Shot examples were provided for simple and complex tasks respectively. The best result is bolded for each task. Also, the performance change from the zero-shot evaluation result is provided in parentheses.

|  | MIMIC-CXR | | IU X-ray | |
| --- | --- | --- | --- | --- |
| Model | SIMPLE (F1) | COMPLEX (F1) | SIMPLE (F1) | COMPLEX (F1) |
| LLaVA-1.5-13B | 9.32(+9.32) | **13.19**(+13.19) | 33.33(+33.33) | 14.04(+14.04) |
| LLaVA-0 | 8.53(-4.38) | 3.36(-2.15) | 38.16(+29.55) | 0.00(+0.00) |
| LLaVA-Med | 3.04(-30.48) | 0.20(-0.20) | 19.91(-9.73) | 0.74(-4.91) |
| LLaVA-1.5-7B | 40.10(+4.31) | 0.00(+0.00) | **47.39**(+6.79) | 4.53(+4.53) |
| LLaVA-CXR-A | 0.00(-39.16) | 7.47(-3.80) | 7.04(-25.32) | **14.05**(-1.61) |
| LLaVA-CXR | **83.29**(+0.11) | 0.00(+0.00) | 40.89(-25.11) | 0.00(+0.00) |

The effect of in-context learning, which is known to improve the performance of a LLM, is shown in Table 3, where 2-shot and 4-shot learning were conducted for SIMPLE and COMPLEX respectively. The performance boost in LLaVA-0 for IU X-ray at SIMPLE difficulty was the highest, +29.55, while the performance decrease in LLaVA-Med for MIMIC-CXR at SIMPLE difficulty was the largest, -30.48. LLaVA-CXR also suffered from a performance decrease in the IU X-ray SIMPLE task. Table 3 shows that LLaVA-1.5-7B and 13B are the only models where providing examples of each choice does not decrease the performance of the model in all difficulties and datasets. The LLaVA-1.5-13B model gained the most performance increase with in-context learning. On the other hand, LLaVA-Med was the only model that performed worse with in-context learning in all difficulties and datasets.

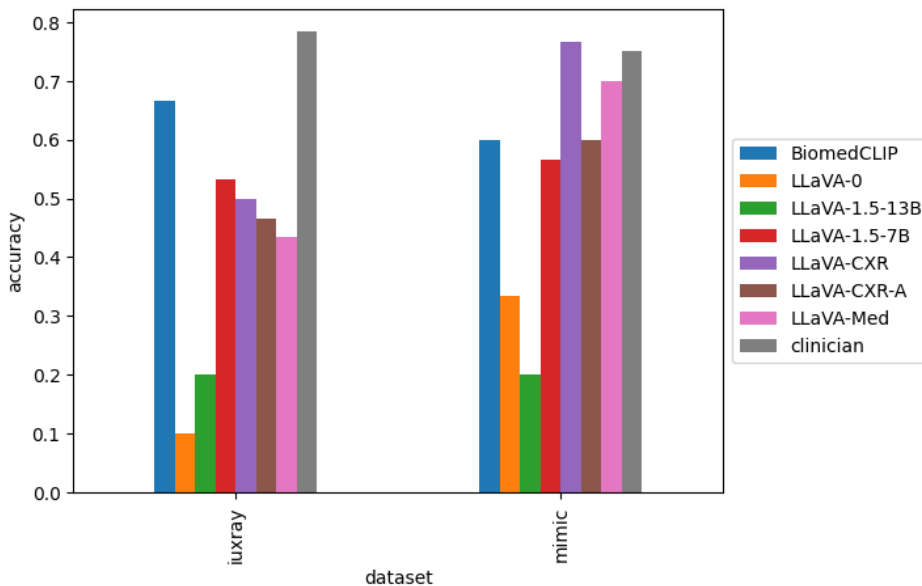

**Figure 5.** Accuracy of human evaluation in the SIMPLE setting.

**Human Evaluation Result**

Figure 5 shows the accuracy of the human evaluation for all the models (zero-shot) and clinicians for the SIMPLE difficulty. The domain experts performed best in IU X-ray with 78.3% accuracy. In the order of accuracy, BiomedCLIP, LLaVA-1.5-7B, LLaVA-CXR, LLaVA-CXR-A, LLaVA-Med, LLaVA-1.5-13B, and LLaVA-0 performed 66.7%, 53.3%, 50.0%, 46.7%, 43.3%, 20.0%, and 10.0% respectively. The accuracy gap between the best model, BiomedCLIP, and clinicians was 11.6%. However, for the MIMIC-CXR dataset, LLaVA-CXR outperformed with an accuracy of 76.7%, surpassing domain experts at 75.0%. LLaVA-Med was slightly worse than domain experts with 70.0% accuracy. LLaVA-CXR-A and BiomedCLIP performed 60.0%, while LLaVA-1.5-7B and LLaVA-0 performed relatively poorly, 56.7% and 33.3%, respectively. This result highlights the effect of fine-tuning VLMs in the target domain.

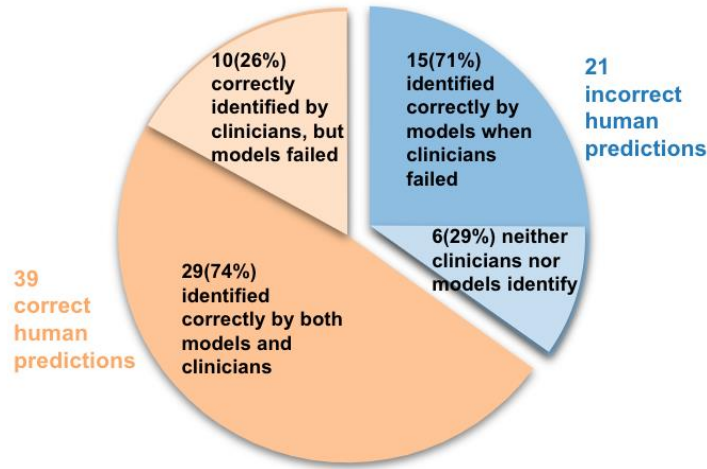

**Figure 6.** Vision LLMs can assist clinicians. The pie chart illustrates the human performance vs ensemble model performance. In 39 (65%) of the reports, both clinicians predicted correctly, whereas in 21 (35%) reports, one of the clinicians failed.

Further analysis of the human evaluation dataset reveals interesting findings indicating the feasibility of Vision LLMs as clinical assistants, as depicted in Figure 6. For this analysis, we used an ensemble of LLaVA models, hereafter referred to as **LLaVA-ensemble**, with LLaVA-1.5-7B, LLaVA-Med, and LLaVA-CXR to make an odd number of voting models. We used majority voting as well as logical OR operation to aggregate the responses from the three models. Figure 6 shows the result of the logical OR operation ensemble. In the total 60 examples, there was a subset of 21 samples (the big circle on the left in Figure 6), for each of which at least one domain expert failed to get the right answer. For this subset, **LLaVA-ensemble** was able to correctly check 15 (71.4%) reports. However, the majority voting aggregation had lower correct reports, 8 (38.1%) reports.

Figure 6 also demonstrates the performance of the models on the subset(n=39) in which both clinicians got the correct answers. Out of these reports, LLaVA-ensemble correctly identified the existence of the mistakes in 29 reports (74.4%) regardless of the aggregation policy (voting and logical OR).

These results underscore the synergistic potential of LLaVA-ensemble as a clinical decision support tool, capable of rectifying oversights made by radiologists. The ensemble's ability to correctly identify the reports where errors eluded domain expert detection highlights its complementary role in enhancing diagnostic accuracy. Moreover, the comparison between aggregation techniques reveals that logical OR operation, by considering a positive prediction if at least one model responds positively, captures subtle errors that can be overlooked.

**DISCUSSION**

In exploring the usefulness of multimodal LLMs in enhancing radiology report error-checking, our study reveals interesting findings across various evaluation approaches. The zero-shot evaluation results demonstrate the effectiveness of instruction tuning in the medical domain, enhancing the model's innate capacity for error-checking in the SIMPLE setting (Table 2). Notably, LLaVA-CXR emerges as the best-performing model for SIMPLE difficulty, highlighting proficiency in understanding radiographs without additional examples. The effectiveness of instruction tuning is also confirmed when we compare the result of LLaVA-Med and its baseline model LLaVA-0. Moreover, our findings on the INSPECT dataset support the robustness and effectiveness of our model across unseen imaging modalities. We observed promising results, further strengthening the applicability of our proposed approach in CT scans.

However, when the examples are provided for in-context learning, both fine-tuned models performed relatively poorly compared to their baseline open domain models (Table 3). LLaVA-1.5-13B emerges as the most robust model, where providing examples for each choice consistently enhances performance. Among the 7B models, LLaVA-1.5-7B was the only one that showed no decrease in performance with additional examples. The difference between the results of the LLaVA-1.5-13B and 7B models suggests that the model size may be the limiting factor of the poor performance in 7B models. The LLaVA-0-7B model also showed a slight decrease in performance in some tasks, while it gained a significant performance boost (29.55%) in the IU X-ray SIMPLE setting. For domain-adapted models, a consistent performance decrease was observed with in-context learning for all models. This is possibly associated with the LLaVA model's limitation with in-context learning that is further exacerbated in the domain adaptation process. However, further investigation is required to determine whether fine-tuning in the medical domain truly limits the model capacity of in-context learning for smaller models.

Additionally, the LLaVA models' architecture does not support the incorporation of multiple input images alongside text prompts for in-context learning. In our experiments, we were constrained to using only textual prompts without the added contextual richness that multiple images could potentially provide. These limitations can significantly impact the VLM's ability to effectively learn from the few-shot examples.

Zero-shot and Few-shot evaluation results highlight the importance of the backbone LLM. LLaVA-1.5-7B outperforms LLaVA-Med (which is based on LLaVA-0) in all cases except the zero-shot COMPLEX setting. This is also confirmed by qualitative analysis (Figure 4). As the LLaVA models tested in this study are all based on a general domain LLM, a future study on VLMs made from a medical LLM could provide insights into the importance of the backbone model.

The evaluation by expert clinicians provides valuable insights. Although the domain experts performed generally better than the models, the LLaVA-CXR model, which was fine-tuned on the MIMIC-CXR instruction dataset for error-checking, performed 1.67% better than humans in the MIMIC-CXR dataset. The detailed analysis of the human evaluation dataset also suggests the potential of Vision LLMs as valuable assistants in clinical settings. LLaVA-ensemble can correct mistakes overlooked by domain experts, showcasing the potential of this model as an assistant in clinical decision-making. Still, the fine-tuned model performed worse than humans with the unseen data format, the IU X-ray dataset, suggesting the need to add more public and custom data to fine-tune the model before it is implemented in hospital settings.

In addition, we recognize that our dataset may not fully replicate the complexities of real-world scenarios. A notable instance is the deidentification issue within the report which confuses both the model and clinicians, for example, 'XXXX' is used to mask private information. This issue occasionally led to confusion for both the model and clinicians during decision-making as it could be difficult to fully exclude a 'substitute' error. This issue was particularly prominent in the IU X-ray dataset, potentially contributing to the observed variance in performance between the IU X-ray and MIMIC-CXR datasets in certain scenarios.

Clinicians observed that some synthetic errors, especially the 'insert' and 'substitute' types, appeared somewhat unrealistic in the evaluation dataset. In some cases, the errors seemed unnatural in real-world reports, except in rare voice dictation errors. Thus, these results may be particularly useful for voice recognition/dictation software errors in radiology reports.

Identifying 'remove' errors presents a challenge due to the inherent variability in the dataset's unstructured radiology reports. In this study, 'remove' errors were marked only when omitted information was considered crucial by the evaluator e.g. when a significant pneumothorax was not reported. However, this approach is inherently subjective, and what is deemed a non-error in one instance might be argued otherwise by another, especially in the absence of structured reporting guidelines. This highlights the ongoing issue of subjectivity in error identification within unstructured radiological assessments.

Furthermore, radiologists typically consider a range of contextual and clinical information, including input from colleagues and previous imaging studies, to form a comprehensive judgment of an image. Our evaluation process was limited in this regard, as it did not fully incorporate these additional data sources.

In future studies, it would be valuable to explore more on enhancing the explanatory capabilities of models analyzing medical images. For example, to not only identify whether it has a mistake but also to provide detailed explanations of their interpretations. This would deepen our understanding of how these models perceive and interpret medical images, advancing the field of AI in diagnostic radiology.

**CONCLUSION**

In conclusion, this research highlights the potential groundbreaking role of multimodal LLMs in radiology, emphasizing their value as assistants for radiologists. The study's thorough evaluation and analysis, encompassing zero-shot, fine-tuning, and in-context learning, alongside human assessments, provide a comprehensive understanding of these models in radiology report error-checking. It underscores the strengths, limitations, and practical implications of the current multimodal LLMs, paving the way for ongoing advancements. This research also created an error-checking benchmark dataset, the first of its kind. This will facilitate the research of a collaborative framework between AI models and medical professionals for enhancing diagnostic precision and operational efficiency in healthcare, thereby opening new avenues for AI-assisted medical imaging.

# Appendix

**Table 4.** Ablation study for In-context learning with LLaVA-1.5-7B. We repeated the experiment 10 times. The reported F1 score is the average of the 10 repeats, and the standard deviation is provided.

|  | MIMIC-CXR | | IU X-ray | |
| --- | --- | --- | --- | --- |
| # Shots | SIMPLE(F1) | COMPLEX(F1) | SIMPLE(F1) | COMPLEX(F1) |
| 0 | 35.79±0.00 | 0.00±0.00 | 40.60±0.00 | 0.00±0.00 |
| 1 | 39.29±1.47 | 9.75±4.48 | 44.22±2.96 | 13.37±6.04 |
| 2 | **39.78**±1.27 | 13.98±5.75 | 39.55±8.62 | 12.81±6.02 |
| 3 | 35.04±9.45 | **14.35**±4.86 | 41.74±2.49 | 9.64±7.04 |
| 4 | 25.22±8.76 | 8.70±9.55 | 42.04±4.56 | 9.74±5.57 |
| 5 | 34.13±12.82 | 10.31±8.70 | 44.03±6.15 | 8.87±5.40 |
| 6 | 34.24±12.07 | 9.07±8.05 | 44.02±4.98 | 8.91±5.87 |
| 7 | 35.83±9.93 | 7.52±6.65 | 44.24±2.19 | 11.15±7.81 |
| 8 | 37.69±9.67 | 9.22±7.68 | 43.81±2.48 | **13.93**±8.71 |
| 9 | 35.64±10.47 | 7.87±5.92 | **44.50**±2.54 | 11.07±7.68 |
| 10 | 34.13±12.34 | 7.33±7.04 | 41.94±4.04 | 12.00±8.30 |

We conducted an ablation study to identify the optimal learning capacity with varying numbers of shots from 0 to 10. Table 4 shows the result of the ablation study for in-context learning with LLaVA-1.5-7B. We repeated the experiment 10 times. The result shows that with in-context learning, the performance slightly improved for both MIMIC-CXR and IU X-ray with both COMPLEX and SIMPLE difficulties, however, there is no significant pattern on the relationship between increase of the few-shots and the performance. Moreover, we notice that the effect of in-context learning is higher in COMPLEX than in SIMPLE difficulty. Providing one example increases the performance in MIMIC-CXR SIMPLE by 3.5 and in IU X-ray SIMPLE by 3.62, while it increases the performance in COMPLEX setting in MIMIC-CXR and IU X-ray by 9.75 and 13.37 respectively. This indicates its potential for addressing more intricate tasks.

**Table 5.** Important clinical concepts annotated by clinicians are randomly modified in the evaluation data.

| Concept | Semantic type |
| --- | --- |
| Effusion pleural | Disease or Syndrome |
| Consolidation | Disease or Syndrome |
| Opacity | Finding |
| Atelectases | Pathologic Function |
| Wet Lung | Pathologic Function |
| Effusion | Pathologic Function |
| No gross lesions | Finding |
| Pneumonia | Disease or Syndrome |
| Congestion | Pathologic Function |
| Dropsy | Pathologic Function |
| Fracture rib | Injury or Poisoning |
| Opaque | Finding |
| Left lung base | Body Location or Region |
| Right lung base | Body Location or Region |
| Right lower lobe | Body Part, Organ, or Organ Component |
| Infection | Disease or Syndrome |
| Fracture | Injury or Poisoning |
| Thoracic aorta | Body Part, Organ, or Organ Component |
| Lobe | Body Part, Organ, or Organ Component |
| COPD NOS | Disease or Syndrome |
| Emphysema | Pathologic Function |
| Displaced fracture | Injury or Poisoning |
| H hernia | Disease or Syndrome |
| Pulmonary edema interstitial | Disease or Syndrome |
| right-sided pleural effusion | Pathologic Function |
| Both lungs | Body Part, Organ, or Organ Component |
| Drainage | Therapeutic or Preventive Procedure |
| left-sided pleural effusion | Pathologic Function |
| Infectious process | Pathologic Function |
| Pulmonary nodules | Finding |
| Lung nodule | Finding |
| Peribronchial cuffing | Finding |
| Right pneumothorax | Finding |